\documentclass[10pt,twocolumn,letterpaper]{article}

\usepackage[pagenumbers]{cvpr} 

\definecolor{cvprblue}{rgb}{0.21,0.49,0.74}
\usepackage[pagebackref,breaklinks,colorlinks,allcolors=cvprblue]{hyperref}

\def\paperID{7213} 
\def\confName{CVPR}
\def\confYear{2026}

\title{Beyond Success: Refining Elegant Robot Manipulation from Mixed-Quality Data via Just-in-Time Intervention}

\author{
    Yanbo Mao$^1$ \quad Jianlong Fu$^2$ \quad Ruoxuan Zhang$^1$ \quad Hongxia Xie$^1$ \quad Meibao Yao$^1$ \\
    $^1$Jilin University \qquad
    $^2$Microsoft Research Asia\\
}

\begin{document}

 \twocolumn[{%
\renewcommand\twocolumn[1][]{#1}%
\maketitle
\begin{center}
    \centering
    \includegraphics[width=1\textwidth,height=8.5cm]{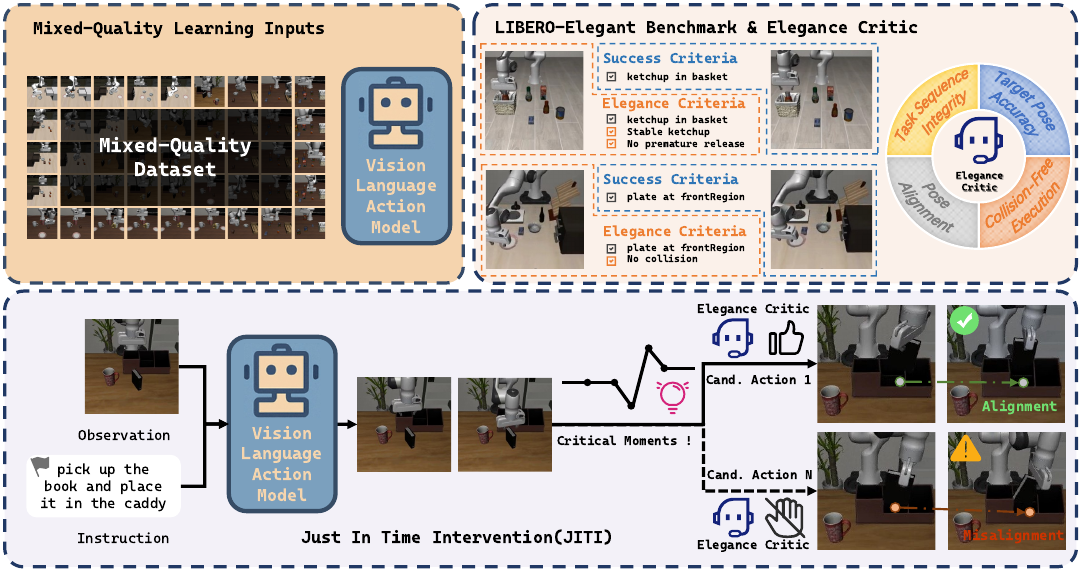}
    \captionof{figure}{\textbf{Overview of our motivation, benchmark, and method.} 
    (Top left) Due to the mixed-quality nature of human demonstrations, VLA models exhibit correspondingly mixed-quality execution at inference time.
    (Top right) The proposed LIBERO-Elegant Benchmark introduces explicit Success and Elegance Criteria, enabling an Elegance-Enriched Dataset and the training of an Elegance Critic for assessing execution quality.
    (Bottom) At inference time, our Just-in-Time Intervention (JITI) mechanism monitors the critic’s confidence and selectively refines the policy at critical moments, providing non-invasive, value-guided improvements in execution elegance.
    }
    \label{fig:teaser}
\end{center}%
}]

\maketitle

\begin{abstract}
Vision-Language-Action (VLA) models have enabled notable progress in general-purpose robotic manipulation, yet their learned policies often exhibit variable execution quality. We attribute this variability to the mixed-quality nature of human demonstrations, where the implicit principles that govern how actions should be carried out are only partially satisfied.
To address this challenge, we introduce the LIBERO-Elegant benchmark with explicit criteria for evaluating execution quality. Using these criteria, we develop a decoupled refinement framework that improves execution quality without modifying or retraining the base VLA policy.
We formalize Elegant Execution as the satisfaction of Implicit Task Constraints (ITCs) and train an Elegance Critic via offline Calibrated Q-Learning to estimate the expected quality of candidate actions. At inference time, a Just-in-Time Intervention (JITI) mechanism monitors critic confidence and intervenes only at decision-critical moments, providing selective, on-demand refinement. 
Experiments on LIBERO-Elegant and real-world manipulation tasks show that the learned Elegance Critic substantially improves execution quality, even on unseen tasks.
The proposed model enables robotic control that values not only whether tasks succeed, but also how they are performed.
\end{abstract}
\section{Introduction}
\label{sec:intro}

A long-standing goal in robotics is to deploy autonomous agents that can operate with human-like dexterity and safety in everyday environments. 
Imitation learning, especially in the form of large-scale Vision–Language–Action (VLA) models~\cite{zhou2025physvlm, huang2025roboground, song2025robospatial, miao2025fedvla, pi0, gr00tn1_2025}, has become a leading paradigm for realizing this vision.
Trained on extensive, internet-scale demonstrations, these models can interpret natural-language instructions and generalize to novel scenes, marking a substantial step forward in general-purpose robotic intelligence~\cite{zhang2025vlabench}.

However, the performance of these powerful models is fundamentally tied to the quality of their training data, revealing a critical yet often underestimated challenge. Real-world demonstration data is inherently heterogeneous and contains demonstrations of mixed and often sub-optimal quality \cite{Mu_2025_CVPR, li2024evaluating}.
It is a noisy mixture of expert-level executions, hesitant corrections, inefficient movements, and even outright failures. 
Standard behavioral cloning inherits this full behavioral distribution, resulting in policies that can produce precise, expert-like motions, but may also express suboptimal or inconsistent behaviors at inference time. The capability for high-quality execution exists within the model, yet it is not expressed reliably.
For instance, in a placement task, the policy may either lower an object to a stable, well-aligned contact or release it too early, causing it to fall or bounce on the target surface.

To identify what differentiates high-quality behaviors from flawed ones, we we draw on the notion of \textbf{Implicit Task Constraints (ITCs)}, the latent rules that specify how an action should be executed, beyond merely achieving the final outcome.
These constraints include proper timing of releases, precise placement, pose alignment, and avoiding unintended contacts. 
We refer to trajectories that satisfy these constraints while also completing the task as \textbf{Elegant Executions}.
To make these implicit constraints operational, we introduce the LIBERO-Elegant Benchmark, which provides a standardized set of manipulation tasks for studying execution quality in a controlled setting.
The key question then is how a robot can learn to infer and fulfill these constraints from demonstrations where they are only intermittently satisfied.

In seeking an answer, we turn to how humans refine their own motor skills. When practicing a task, humans do not correct their motions uniformly along the entire trajectory; instead, they selectively attend to a small number of \emph{decision-critical moments} where subtle adjustments have disproportionate influence on the final outcome. 
This insight suggests a similar decomposition for robots: evaluate while executing.
A pre-trained VLA policy handles broad task execution, while a lightweight \emph{Elegance Critic} is responsible solely for assessing execution quality. During inference, our \emph{Just-in-Time Intervention (JITI)} mechanism monitors the critic’s uncertainty and intervenes only at decision-critical states, providing selective, non-invasive refinement rather than continuous control (see Fig.~\ref{fig:teaser}).

Building on these observations, this work makes the following contributions:
\begin{itemize}
    \item We formalize \textbf{Elegant Execution} as the reliable satisfaction of \textbf{Implicit Task Constraints} (ITCs), shifting the focus from binary success to the quality of manipulation.
    \item We construct the \textbf{LIBERO-Elegant Benchmark}, a curated suite of tasks that evaluates motion quality beyond binary success, enabling systematic study of execution quality in manipulation.
    \item We propose a \emph{non-invasive refinement framework} that decouples execution from evaluation through an \textbf{Elegance Critic} and a \textbf{Just-in-Time Intervention (JITI)} mechanism.
    \item Through comprehensive experiments in both simulation and real-world settings, we demonstrate significant gains in execution quality while further confirming the effectiveness and robust generalization of our approach across diverse manipulation tasks.
\end{itemize}

\section{Related Works}
\label{sec:related_works}

\noindent \textbf{Data-Centric Approaches to Imitation Learning.} Data-Centric Approaches address the challenge of mixed-quality data by operating directly on the dataset, optimizing the policy via purification or restructuring. The core hypothesis for this class of methods is that standard Behavioral Cloning (BC) will yield a superior policy, provided that the data distribution it imitates is improved. Common techniques include training discriminators or self-supervised models to filter or re-weight data \cite{DWIL, Tangkaratt2019VILDVI, brown2019drex, Wu2024LearningFI}, using importance sampling to correct for distribution shift \cite{ILI}, or leveraging retrieval and conditional generation to extract useful behaviors from unlabeled data \cite{Du2023BehaviorRF, C-BeT, chen2025vidbot}.

However, these methods remain short-sighted: they reshape the static dataset rather than learning value functions that reason about long-term consequences. This motivates the shift toward Reinforcement Learning.

\noindent \textbf{Policy Improvement via Online Reinforcement Learning.} Online Reinforcement Learning (Online RL) is a common strategy for improving policies beyond BC. It typically follows a ``pre-train and fine-tune'' paradigm: a policy, often a large VLA model, is initialized via supervised fine-tuning or BC and then improved through online interaction in a live environment.
This remains a highly active research area, with numerous frameworks proposed to enhance policies beyond their static-data initialization \cite{Shu2025RFTFRF, Chen2025ConRFTAR, Lu2025VLARLTM, Chen2025TGRPOV, Tan2025InteractivePF, Ren2024DiffusionPP, Guo2025ImprovingVM, Luo2024PreciseAD, xu2024rldgroboticgeneralistpolicy, liu2025can, mark2024policy}.

However, Online RL poses major challenges for large-scale robotic VLAs: it demands costly and unsafe real-world interaction, and fine-tuning remains invasive and unstable. These issues motivate a shift toward Offline Reinforcement Learning.

\noindent \textbf{Policy Improvement via Offline Reinforcement Learning.}
Offline Reinforcement Learning (Offline RL) improves policies from static datasets without any online interaction, addressing the cost and safety concerns of the online paradigm. However, it faces the out-of-distribution (OOD) problem, where distributional shift can induce value overestimation. Representative algorithms such as CQL~\cite{kumar2020conservative}, IQL~\cite{kostrikov2021offline}, and Cal-QL~\cite{nakamoto2023cal} mitigate this via conservative objectives and value calibration that constrain learning to the dataset’s support.

Building on these foundations, recent studies have explored integrating Offline RL with large VLA models for data-efficient refinement. Two paradigms have emerged: (i) invasive policy optimization, which fine-tunes pretrained VLA models ~\cite{LWID, Zhang2025ReinboTAR, zolna2020offline} but risks catastrophic forgetting, and (ii) decoupled inference-time guidance, which preserves the base model and employs a lightweight critic to guide outputs during inference~\cite{song2025hume, nakamoto2024steering, xia2025phoenix, zhou2025code}. The latter is non-invasive and retains the generality of pretrained policies.

Our work situates firmly within this decoupled paradigm, but pursues a more quality-aware objective: \textit{elegant execution}. It is defined as the satisfaction of fine-grained, implicit task constraints often overlooked in mixed-quality demonstrations.
To this end, we introduce the LIBERO-Elegant Benchmark and train a lightweight critic to deliver just-in-time, non-invasive guidance during inference.

\section{LIBERO-Elegant Benchmark}
\label{sec:libero_elegant}

\subsection{Overview}
To evaluate manipulation quality beyond binary task success, we introduce the LIBERO-Elegant Benchmark, a curated extension of LIBERO~\cite{liu2023libero}. Although every demonstration in LIBERO is labeled as successful, they differ substantially in how implicit task constraints (ITCs) are satisfied. LIBERO-Elegant makes this variation measurable through a set of quality-sensitive tasks, unified elegance criteria, and an \textit{Elegance-Enriched Dataset} used for training value-based critics.

\subsection{Task Design and Evaluation Criteria}
LIBERO-Elegant contains eight manipulation tasks selected for their sensitivity to execution quality. 
Typical behaviors include precise placement, controlled insertion, and collision-free pushing, each governed by implicit constraints such as accurate centering, stable contact, and obstacle avoidance.

Each task is evaluated under two complementary criteria:
\textbf{Success Criteria}, which follow LIBERO’s original goal conditions, 
and \textbf{Elegance Criteria}, which assess execution quality along four dimensions:
\begin{itemize}[leftmargin=*]
\item \textbf{Task Sequence Integrity}: whether the action sequence follows the intended steps and completes each required phase.
\item \textbf{Target Pose Accuracy}: whether the final object position lies within the required spatial tolerance.
\item \textbf{Pose Alignment}: whether the final orientation matches the task-specific rotational target.
\item \textbf{Collision-Free Execution}: whether the motion avoids unintended contact with the environment or other objects.
\end{itemize}
These dimensions provide a consistent framework for evaluating manipulation elegance across tasks. 
Additional task details are provided in Supp.~\ref{sec:libero_tasklist}.

\subsection{Annotation and Reward Design}
To provide supervision for evaluating elegance, we annotate short temporal segments where ITCs are most relevant and assign binary rewards $r_t \in \{0,1\}$ indicating whether the constraints are satisfied. Premature releases or misaligned placements receive $0$, while controlled, constraint-satisfying transitions receive $1$. The resulting \textit{Elegance-Enriched Dataset} $\mathcal{D}_\text{elegant}$ augments each demonstration with a short labeled segment indicating whether the implicit constraints are satisfied. All annotations are cross-validated by multiple raters for consistency, with further implementation details in Supp.~\ref{sec:libero_annotation}.

\subsection{Data Statistics}
LIBERO-Elegant consists of \textbf{8 manipulation tasks}, covering \textbf{327 demonstration episodes} and approximately \textbf{52.7K frames} of synchronized RGB-D and proprioceptive data. Among them, \textbf{148 episodes} exhibit high-quality execution and receive positive elegance rewards. Each rewarded episode is annotated over a single 25-frame window corresponding to the key moment where ITCs are most critical.

\section{Method}
\label{sec:method}

\begin{figure*}[t]
    \centering 
    \includegraphics[width=\textwidth]{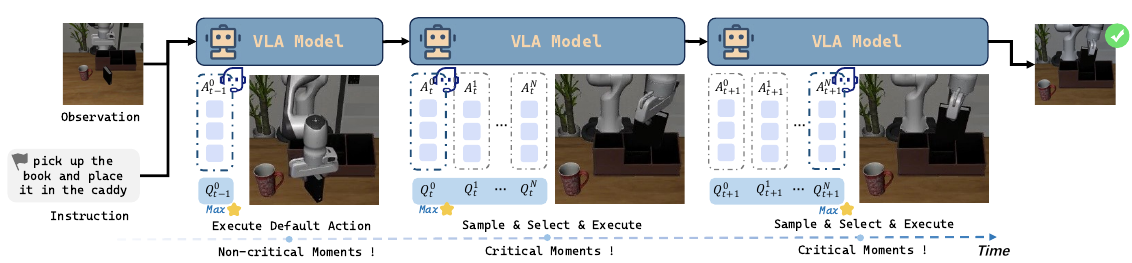}
    \caption{\textbf{Illustration of the proposed Just-in-Time Intervention (JITI) process.} At each timestep, the Elegance Critic monitors its predicted value. When confidence remains stable (non-critical moment), the default action from the base policy is executed directly. When significant value fluctuation is detected, JITI triggers multi-sample evaluation and selects the action with the highest predicted elegance.}
    \label{fig:method_overview}
\end{figure*}

\label{sec:method}

\begin{figure}[t]
    \centering 
    \includegraphics[width=0.45\textwidth]{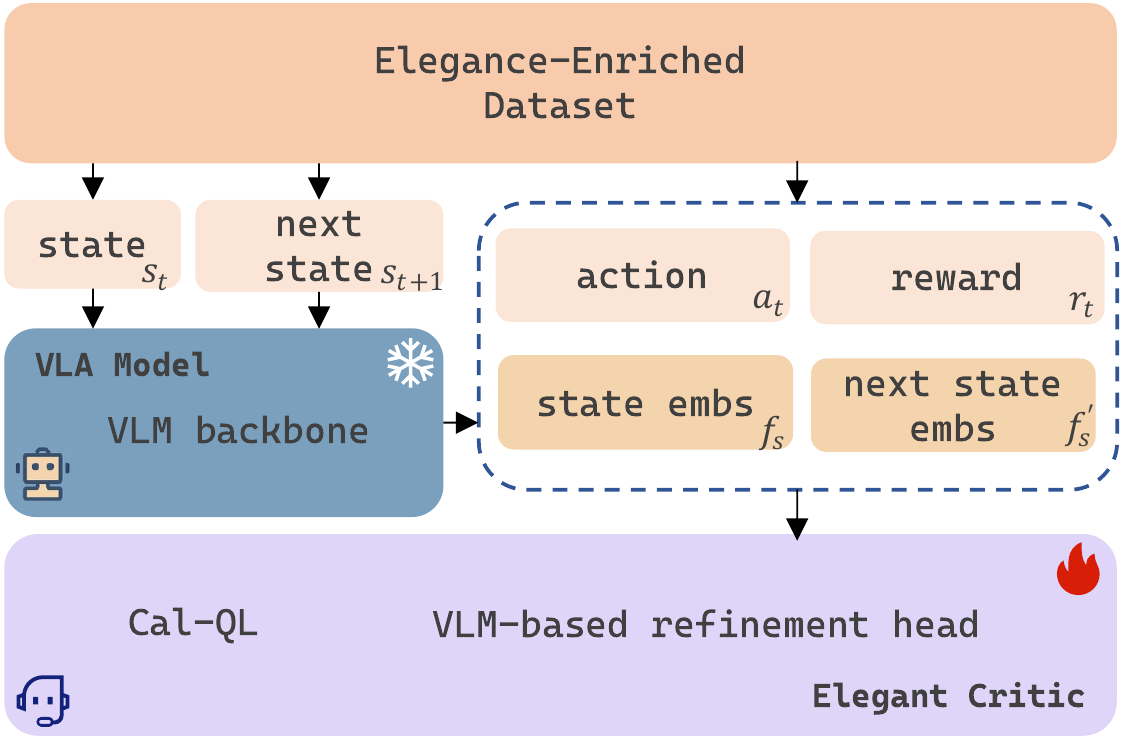}
    \caption{\textbf{Training pipeline of the Elegance Critic.} Samples from the Elegance-Enriched Dataset are first processed by the frozen VLA backbone to obtain contextual embeddings. These embeddings, together with the corresponding actions and graded rewards, are then fed into the Calibrated Q-Learning (Cal-QL) module, which refines them to update the critic.}
    \label{fig:network_architecture}
\end{figure}

Our goal is to enhance the elegance of robotic manipulation by encouraging behaviors that better satisfy \emph{Implicit Task Constraints} (ITCs). 
To this end, we adopt a decoupled three-stage framework (overview in \cref{fig:method_overview}) that performs evaluation during execution. 
A base Vision-Language-Action (VLA) policy executes the task, while a lightweight \emph{Elegance Critic} assesses motion quality and provides selective, inference-time guidance.

\textbf{Stage~1} trains a base VLA policy $\pi_\theta$ on mixed-quality demonstrations to capture the overall behavioral distribution (\cref{sec:stage_one}).  
\textbf{Stage~2} uses the LIBERO-Elegant annotations to train an Elegance Critic $Q_\phi$ via Calibrated Q-Learning (Cal-QL)~\cite{nakamoto2023cal} (\cref{sec:stage_two}).  
\textbf{Stage~3} integrates the two through the \emph{Just-in-Time Intervention (JITI)} mechanism, which monitors critic confidence and refines actions only at decision-critical moments, without retraining or continuous guidance (\cref{sec:stage_three}).




\subsection{Base Generative Policy Training}
\label{sec:stage_one}

We begin \textbf{Stage~1} by training a base generative policy $\pi_\theta$ designed to capture the full distribution of human actions within the mixed-quality dataset. Rather than differentiating between good and poor demonstrations at this stage, $\pi_\theta$ is trained to model the overall behavioral distribution $p(A_t \mid s_t)$ observed in human data.

The policy $\pi_\theta$ is implemented as a flow-matching generative model \cite{Lipman2022FlowMF}. At its core lies a Transformer-based network $v_\theta$, which learns a continuous-time vector field that transforms a noise sample into a clean action. 
Specifically, the network $v_\theta(A_t^\tau, s_t)$ takes the current state $s_t$ together with a ``noisy action'' $A_t^\tau$ corresponding to a noise level $\tau \in [0, 1]$ as input. It then predicts the instantaneous flow direction that moves $A_t^\tau$ toward the target action $A_t$.

During training, we sample a ground-truth action $A_t$ from the mixed-quality dataset $\mathcal{D}$ and a Gaussian noise vector $\epsilon \sim \mathcal{N}(0, I)$. A noisy action is constructed as $A_t^\tau = \tau A_t + (1-\tau)\epsilon$ for a random $\tau$. The network parameters $\theta$ are then optimized by minimizing the mean squared error between the predicted and target vector fields:

\begin{equation}
\mathcal{L}_{\text{FM}}(\theta) = 
\mathbb{E}_{\tau, (s_t, A_t), \, \epsilon} 
\left[
\left\| v_\theta(A_t^\tau, s_t) - (A_t - \epsilon) \right\|_2^2
\right].
\label{eq:flow_matching_loss}
\end{equation}

This objective trains $v_\theta$ to estimate the direction from a noisy sample $\epsilon$ toward its corresponding clean action $A_t$, enabling the model to learn a smooth mapping between noise and action space.

After training, the policy $\pi_\theta$ functions as a generative model that can sample actions conditioned on a given state $s_t$. 
Starting from an initial noise vector $A_t^0 \sim \mathcal{N}(0, I)$, the model integrates the learned vector field $v_\theta$ over $\tau \in [0, 1]$. A numerical solver, such as the forward Euler method, is then applied to obtain the clean action sample.


Importantly, by initializing with multiple noise samples, $\pi_\theta$ can generate a diverse set of candidate actions for the same state $s_t$. This diversity provides the foundation for the value-guided selection mechanism used in our Just-in-Time Intervention (JITI) algorithm introduced in Stage~3.

\subsection{Offline Elegance Critic Training}
\label{sec:stage_two}

In \textbf{Stage~2}, we train an Elegance Critic $Q_\phi(s_t, a_t)$, which provides an evaluative signal for selecting elegant actions during inference (Stage~3).  
The critic is trained to estimate the expected cumulative return that reflects both task success and behavioral elegance, leveraging the graded reward annotations described in Sec.~\ref{sec:libero_elegant}.  
This allows it to distinguish subtle differences in motion quality that are not captured by binary success labels.

\noindent \textbf{Elegance Critic Architecture.}
The training pipeline of the Elegance Critic is illustrated in Fig.~\ref{fig:network_architecture}. 
During training, the critic learns from the \textit{Elegance-Enriched Dataset} $\mathcal{D}_\text{elegant}$ containing tuples $(s_t, a_t, r_t, s_{t+1})$, 
where each reward $r_t$ encodes the degree of elegance according to implicit task constraints.
To leverage prior knowledge while remaining modular, the critic builds upon the pre-trained VLA foundation model from Stage~1 in a decoupled manner.
The multi-modal input states $s_t$ and $s_{t+1}$, comprising visual observations, language instruction, and robot proprioception, are passed through the frozen VLM backbone to extract intermediate representations.
These features are passed to a VLM-based refinement head, which repurposes the frozen backbone representations for value estimation. This allows the critic to capture task- and reward-relevant nuances without modifying the pre-trained encoder.
The resulting contextual embeddings $f_s$ and $f_{s'}$ are concatenated with the action $a_t$ and the annotated reward $r_t$. These features are then passed through the Calibrated Q-Learning (Cal-QL) module to update the critic parameters $\phi$, resulting in the learned value function $Q_\phi(s_t, a_t)$.

\noindent \textbf{Calibrated Q-Learning for Elegance-Aware Value Estimation.}
Learning to evaluate \textit{elegance} from mixed-quality demonstrations poses a dual challenge: the critic must remain sensitive to graded, fine-grained rewards that encode implicit task constraints while avoiding overestimation on unseen or low-quality actions that fall outside the dataset’s support.

To balance \textit{elegance sensitivity} with \textit{distributional conservatism}, we adapt Calibrated Q-Learning (Cal-QL)~\cite{nakamoto2023cal} to train the Elegance Critic. Here, Cal-QL acts as a calibration mechanism that aligns the critic’s value predictions with the behavioral distribution while preserving its responsiveness to the graded elegance rewards.

Specifically, the calibration regularizer is defined as:

\begin{multline}
\mathcal{R}_{\text{cal}}(\phi)
= \mathbb{E}_{s \sim \mathcal{D}} \Big[
\max\!\big(
\mathbb{E}_{a \sim \pi(\cdot|s)} Q_\phi(s, a),\;
V_\mu(s)
\big) \\
-\, \mathbb{E}_{a \sim \mathcal{D}(\cdot|s)} Q_\phi(s, a)
\Big].
\label{eq:calql_reg}
\end{multline}

This formulation ensures that when the critic’s estimate for in-distribution actions is already below
the behavioral value $V_{\mu}(s)$, it does not push them down any further. Thus, the critic’s confidence is properly calibrated, yielding conservative yet accurate value estimates.

The full critic objective is therefore given by:
\begin{equation}
\mathcal{L}_{\text{Cal-QL}}(\phi)
= \mathcal{L}_{\text{Bellman}}(\phi)
+ \lambda_{\text{cal}} \, \mathcal{R}_{\text{cal}}(\phi),
\label{eq:calql_total_loss}
\end{equation}
where $\mathcal{L}_{\text{Bellman}}(\phi)$ enforces temporal consistency:
\begin{multline}
\mathcal{L}_{\text{Bellman}}(\phi)
= \mathbb{E}_{(s_t, a_t, r_t, s_{t+1}) \sim \mathcal{D}}
\Big[
Q_\phi(s_t, a_t) \\
- \big(r_t + \gamma \max_{a'} Q_{\phi'}(s_{t+1}, a')\big)
\Big]^2,
\label{eq:bellman_loss}
\end{multline}
and $Q_{\phi'}$ denotes a slowly updated target network.

Through this formulation, the Elegance Critic learns calibrated value estimates that faithfully capture the nuanced notion of elegance within the data’s support, while maintaining conservatism for unseen behaviors. This balance enables reliable, value-guided refinement during inference (Stage~3).

\subsection{Just-in-Time Intervention (JITI)}
\label{sec:stage_three}

Upon completion of offline training, the framework yields a generative base policy $\pi_\theta$ (Stage~1) capable of stochastically producing trajectories of varying quality, and an Elegance Critic $Q_\phi$ (Stage~2) trained to evaluate the expected elegance of state–action pairs. The final stage integrates these two components at inference time.

A naive solution is to sample multiple trajectories from $\pi_\theta$ at every timestep, evaluate them all using $Q_\phi$, and execute the most elegant action. Although effective, this ``full guidance'' strategy is computationally expensive and unnecessary, most decisions are locally unambiguous. In practice, only a few \emph{critical moments} significantly influence the overall elegance of a trajectory. 

The key insight motivating our method is that a trajectory’s overall elegance is primarily determined by a small number of \emph{critical moments}, instances where local decisions have a disproportionate influence on the global quality of the motion. To exploit this property, we propose the \textbf{Just-in-Time Intervention (JITI)} algorithm: an event-driven, plug-and-play guidance mechanism that activates high-cost multi-sample evaluation only when the policy exhibits signs of uncertainty.

\noindent \textbf{Identifying Critical Moments via Q-Value Fluctuation.}
The Elegance Critic $Q_\phi$ serves as a learned estimator of the expected elegance achievable from a given state-action pair $(s_t, a_t)$. Intuitively, when the base policy $\pi_\theta$ operates in familiar regions and proposes actions consistent with the learned notion of \emph{elegance}, the critic’s Q-value predictions remain stable and high-confidence. 

Conversely, when the policy encounters out-of-distribution states or issues a suboptimal action (e.g., an unstable grasp or an inefficient trajectory), its confidence declines.
This instability manifests as a sharp, sudden fluctuation in the predicted Q-values. This fluctuation, either a sudden drop (signaling a loss of value) or a sudden spike (signaling a critical, high-stakes moment), indicates a temporal inconsistency in the critic's evaluation.

These fluctuations naturally arise from the critic’s dual training dynamics. The sharp value spikes reflect \textit{Bellman backups} triggered by the sparse, graded rewards that delineate critical, high-reward temporal segments in the Elegance-Enriched Dataset. The sudden drops stem from the \textit{conservative regularization} in Calibrated Q-Learning (Cal-QL), which penalizes out-of-distribution or non-elegant actions. Together, these effects enable the critic to recognize both \textit{high-value states} and \textit{low-confidence transitions}.

We quantify this behavior via a \emph{Q-value fluctuation metric} $\Delta q_t = |q_t - \bar{q}t|$, where $q_t = Q\phi(s_t, A_t^0)$ is the critic’s evaluation of the default policy action and $\bar{q}_t$ is the moving average over a short history window. 
A small $\Delta q_t$ indicates temporal consistency and high critic confidence.
In contrast, a sudden spike, whether reflecting entry into an ITC-relevant, high-value segment or a drop caused by uncertainty, signals a decision-critical moment.
Thus, $\Delta q_t$ provides a principled, low-cost indicator for when to intervene.

\noindent \textbf{The JITI Guidance Algorithm.}
At inference time, JITI continuously monitors $\Delta q_t$ and compares it to a predefined threshold $\tau$. When $\Delta q_t \le \tau$, the moment is considered non-critical, and the system executes the default policy action $A_t = A_t^0$, requiring only a single critic evaluation. 
When $\Delta q_t > \tau$, JITI identifies a critical moment of heightened uncertainty.
It then intervenes by sampling multiple candidate actions from $\pi_\theta$, scoring them with the Elegance Critic,
and executing the one with the highest predicted elegance.
The complete procedure is summarized in Algorithm~\ref{alg:jiti}.



\begin{algorithm}[t]
\caption{Computation of Q-value Fluctuation and JITI Guidance}
\label{alg:jiti}
\begin{algorithmic}[1]
\Require Base policy $\pi_\theta$, Elegance Critic $Q_\phi$, threshold $\tau$, window size $k$, candidate count $N$
\State Initialize Q-value history buffer $H \gets \emptyset$
\For{each timestep $t = 1, 2, \dots$}
    \State Sample default action $A_t^0 \sim \pi_\theta(s_t)$
    \State Evaluate $q_t = Q_\phi(s_t, A_t^0)$
    \State Update buffer $H_t \gets H_{t-1} \cup \{q_t\}$, keep last $k$ values
    \State Compute mean $\bar{q}_t = \frac{1}{|H_t|}\sum_{q_i \in H_t} q_i$
    \State Compute fluctuation $\Delta q_t = |q_t - \bar{q}_t|$
    \If{$\Delta q_t \le \tau$} 
        \State Execute default action $A_t = A_t^0$
    \Else 
        \State Sample $N$ candidate actions $\mathcal{A}_{\text{c}} = \{A_t^1, \dots, A_t^N\} \sim \pi_\theta(s_t)$
        \State Evaluate each candidate: $q_i = Q_\phi(s_t, A_t^i)$
        \State Select $A_t = \arg\max_{a \in \mathcal{A}_{\text{c}}} Q_\phi(s_t, a)$
        \State Execute $A_t$
    \EndIf
\EndFor
\end{algorithmic}
\end{algorithm}

Through this event-driven guidance scheme, JITI maintains the efficiency of single-sample execution during routine decisions, while selectively invoking multi-sample evaluation at critical points of elevated uncertainty. This mechanism enables \textit{on-demand refinement} without retraining $\pi_\theta$, effectively filtering out sub-optimal behaviors and ensuring consistent elegance across the entire trajectory.

\section{Experiments}
\label{sec:experiments}

We design a series of experiments to systematically answer the following research questions:
\begin{itemize}[leftmargin=*]
    \item \textbf{RQ1: Effectiveness.} Does the proposed JITI-guided Elegance Critic improve task performance and behavioral elegance compared to base vision-language-action (VLA) policies?
    \item \textbf{RQ2: Component Contribution.} How does each module, particularly the Just-in-Time Intervention (JITI) mechanism and the elegance-oriented reward calibration, contribute to the overall performance?
    \item \textbf{RQ3: Generalization.} Can the Elegance Critic generalize its learned concept of ``elegance'' to unseen yet semantically similar manipulation tasks without retraining?
\end{itemize}
Together, these research questions evaluate our method from three key perspectives: \emph{effectiveness}, \emph{modularity}, and \emph{generalization}.

\subsection{Simulation Experimental Setup}
\label{sec:sim_setup}


\noindent \textbf{Benchmark.} All simulation experiments are conducted on the \textbf{LIBERO-Elegant Benchmark} introduced in Sec.~\ref{sec:libero_elegant}, which augments LIBERO tasks with task-specific \emph{elegance criteria} to evaluate execution quality beyond success.

\noindent \textbf{Evaluation Metrics.} Our primary metric is the \textbf{Elegant Success Rate (ESR)}. An episode is considered an elegant success \emph{if and only if} it achieves the task goal \emph{and} satisfies all predefined elegance constraints. All reported ESR values are averaged over 50 evaluation rollouts per task.

\noindent \textbf{Implementation Details.} 
We evaluate two representative Vision-Language-Action (VLA) architectures with different model capacities: \textbf{SmolVLA} (450M) and \textbf{Isaac GR00T N1.5} (3B). 

The SmolVLA base policy is trained for 100k steps (batch size 64) on an NVIDIA RTX 5090 GPU, while the Isaac GR00T N1.5 model is finetuned for 50k steps (batch size 16) on four NVIDIA A40 GPUs following the official GR00T finetuning pipeline.

Both corresponding Elegance Critics are trained for 20k steps (batch size 32) on an RTX 5090 GPU using AdamW with a learning rate of $1 \times 10^{-5}$. And we employ SmolVLM2-256M-Video-Instruct \cite{marafioti2025smolvlm} and Eagle2-1B \cite{li2025eagle} as VLM-based refinement heads for the SmolVLA and GR00T variants, respectively.

\subsection{Effectiveness (RQ1)}
\label{sec:sim_main_results}

We begin by addressing \textbf{RQ1 (Effectiveness)}, whether our JITI-guided Elegance Critic can improve both task success and behavioral elegance over baseline VLA policies. We evaluate this by comparing our method against the unmodified base policies across diverse manipulation tasks in the LIBERO-Elegant benchmark.

As shown in \textbf{Table \ref{tab:main_results}}, our JITI-guided Elegance Critic yields substantial and consistent performance gains across all tasks and both base models.

\begin{table}[h]
\centering
\caption{\textbf{Comprehensive comparison on the LIBERO-Elegant simulation benchmark.} We report Elegant Success Rate (ESR, \%). Full implementation details and hyperparameter settings for all compared methods are provided in Supp.~\ref{sec:supp_imple}.}
\label{tab:main_results}
\begingroup
\fontsize{15}{16}\selectfont
\resizebox{\columnwidth}{!}{%
\begin{tabular}{lcccccccc|c}
\toprule
\textbf{Method} & \textbf{T-0} & \textbf{T-1} & \textbf{T-2} & \textbf{T-3} & \textbf{T-4} & \textbf{T-5} & \textbf{T-6} & \textbf{T-7} & \textbf{Avg.} \\
\midrule
$\pi_{0.5}$\cite{pi05}  & 66 & 24 & 36 & 40 & 38 & 44 & 42 & 64 & 44.2 \\
Isaac GR00T N1 \cite{gr00tn1_2025} & 50 & 42 & 48 & 36 & 36 & 30 & 24 & 56 & 40.2 \\
SmolVLA (Base) \cite{SmolVLA}  & 70 & 30 & 34 & 52 & 50 & 42 & 48 & 72 & 49.8 \\
Isaac GR00T N1.5 (Base) \cite{gr00tn1_2025} & 44 & 22 & 42 & 36 & 36 & 56 & 56 & 76 & 46.0 \\
\midrule
\multicolumn{10}{l}{\textit{Our Method (JITI-Guided)}} \\
\textbf{Ours (JITI) + SmolVLA} & \textbf{86} & \textbf{68} & \textbf{42} & \textbf{62} & \textbf{60} & \textbf{54} & \textbf{74} & \textbf{92} & \textbf{67.2} \\
\textbf{Ours (JITI) + GR00T N1.5} & \textbf{78} & \textbf{36} & \textbf{60} & \textbf{70} & \textbf{62} & \textbf{76} & \textbf{72} & \textbf{84} & \textbf{67.2} \\
\bottomrule
\end{tabular}
}
\endgroup
\end{table}

For \textbf{SmolVLA}, our method improves the average ESR from 49.8\% to 67.2\% (+17.4 pts).
More importantly, our framework demonstrates strong ``plug-and-play'' capability: 
when integrated with the \textbf{GR00T N1.5} model, it achieves a similar improvement from 46.0\% to 67.2\% (+21.2 pts). 
This confirms that our decoupled critic serves as a model-agnostic module capable of enhancing diverse VLA architectures.

\subsection{Ablation Study (RQ2)}
\label{sec:sim_ablation_jiti}

\begin{figure}[t]
    \centering
    \includegraphics[width=0.48\textwidth]{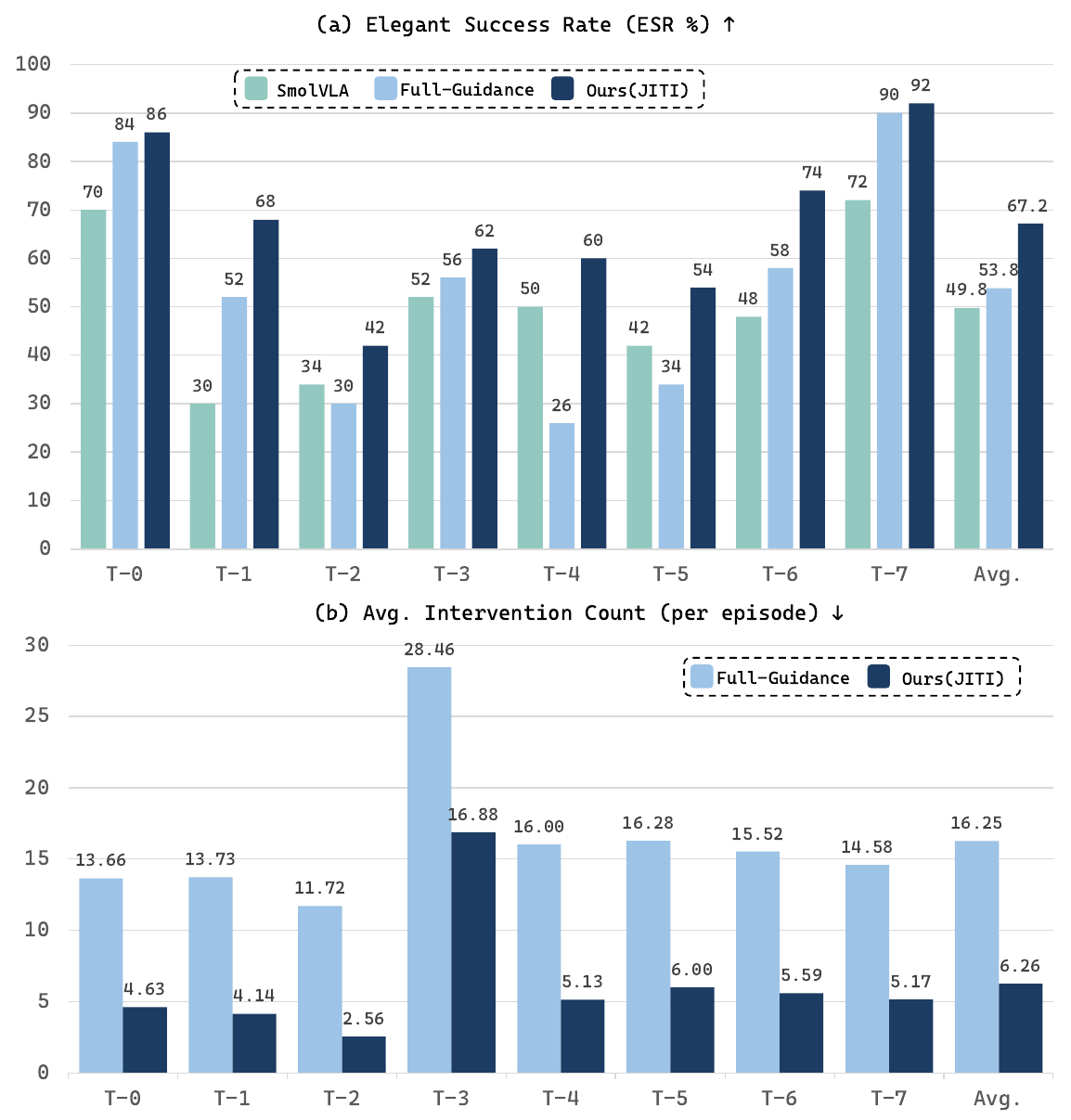}
    \caption{
    \textbf{Ablation on the Just-in-Time Intervention (JITI) mechanism.}
    (a) Elegant Success Rate (ESR, \%) across eight LIBERO-Elegant tasks, comparing the base policy, Full-Guidance, and our JITI-guided variant.
    (b) Average Intervention Count per episode for Full-Guidance and JITI.
    JITI achieves higher ESR with significantly fewer interventions, demonstrating its efficiency and selectivity.
    }
    \label{fig:ablation_study}
\end{figure}

To address \textbf{RQ2: Component Contribution}, we evaluate the effectiveness of the proposed Just-in-Time Intervention (JITI) mechanism by comparing three variants: 
the base policy without guidance, a \texttt{Full-Guidance} variant that applies $N$-sample evaluation at every timestep, 
and our proposed \texttt{JITI} model that triggers guidance only when critic uncertainty rises.

As shown in Fig.~\ref{fig:ablation_study}(a), 
\texttt{JITI} consistently outperforms both baselines in Elegant Success Rate (ESR) across all eight tasks. 
At the same time, Fig.~\ref{fig:ablation_study}(b) shows that it reduces the number of critic interventions by more than 60\% compared to Full-Guidance, 
confirming that targeted, event-driven refinement can preserve efficiency while improving execution quality.

\noindent \textbf{Effect of Elegance-Oriented Reward Calibration.}
Beyond temporal scheduling, we also examine how the elegance-oriented reward shaping contributes to stable and smooth policy behavior.

To evaluate the contribution of our task-specific elegance signal, we train the Elegance Critic under two alternative reward schemes and compare the resulting inference performance. The first scheme (\emph{Binary Reward}) provides only a sparse, end-of-episode success label; the second scheme is our proposed \emph{Task-Specific Reward}, which supplies dense, intermediate supervision that explicitly encodes implicit task constraints.

As shown in Table~\ref{tab:ablation_reward}, the critic trained with the task-specific elegance rewards yields substantially higher ESR (67.2\% vs.\ 56.8\% on average). This gap indicates that sparse binary feedback is insufficient for learning the fine-grained preferences that define elegance: the dense, targeted reward enables the critic to distinguish and prefer subtle, high-quality execution patterns that binary success labels cannot reliably capture. We therefore conclude that effective elegance refinement requires not only an uncertainty-aware intervention mechanism (JITI) but also an appropriately informative training signal for the critic.

\begin{table}[h]
\centering
\caption{\textbf{Ablation on reward formulation in simulation.} The proposed task-specific elegance reward yields higher ESR (\%) $\uparrow$.}
\label{tab:ablation_reward}
\begingroup
\fontsize{15}{16}\selectfont
\resizebox{\columnwidth}{!}{%
\begin{tabular}{lcccccccc|c}
\toprule
\textbf{Reward Type} & \textbf{T-0} & \textbf{T-1} & \textbf{T-2} & \textbf{T-3} & \textbf{T-4} & \textbf{T-5} & \textbf{T-6} & \textbf{T-7} & \textbf{Avg.} \\
\midrule
Binary Reward & 72 & 66 & 34 & 56 & 28 & 44 & 60 & \textbf{94} & 56.8 \\
\textbf{Task-Specific (Ours)} & \textbf{86} & \textbf{68} & \textbf{42} & \textbf{62} & \textbf{60} & \textbf{54} & \textbf{74} & 92 & \textbf{67.2} \\
\bottomrule
\end{tabular}
}
\endgroup
\end{table}

\subsection{Generalization (RQ3)}

\noindent \textbf{Experimental Setup.}
To evaluate whether the Elegance Critic captures a transferable notion of motion quality, we conduct experiments on a set of pick-and-place tasks drawn from the LIBERO-Object suite. 

We split the LIBERO-Object tasks into a seen subset, used to train the critic on objects such as \textit{milk}, \textit{ketchup}, and \textit{alphabet soup}, and an unseen subset containing novel but semantically similar tasks involving objects such as \textit{salad dressing}, \textit{BBQ sauce}, \textit{tomato sauce}, and \textit{orange juice}. 

This setup tests whether the critic generalizes its learned notion of ``elegance'' to unseen objects and spatial contexts while preserving the same high-level pick-and-place semantics. Detailed task specifications and evaluation conditions are provided in Supp.~\ref{sec:supp_rq3}.

\noindent \textbf{Results.}
As shown in Table~\ref{tab:generalization}, our method consistently improves the Elegant Success Rate (ESR) on both seen and unseen tasks. 
Specifically, ESR rises from 54.6\% to 72.0\% on the seen tasks and from 53.0\% to 68.6\% on the unseen ones, demonstrating that the Elegance Critic learns a transferable behavioral prior rather than memorizing task-specific motion patterns.
This enables zero-shot refinement of motion quality across new manipulation contexts.


\begin{table}[h]
\centering
\caption{\textbf{Generalization of the Elegance Critic to unseen tasks.} 
The critic trained on three seen tasks is directly applied to four unseen but semantically similar manipulation tasks without fine-tuning. 
ESR (\%) $\uparrow$}
\label{tab:generalization}
\begingroup
\fontsize{15}{16}\selectfont
\resizebox{\columnwidth}{!}{
\begin{tabular}{lcccccc}
\toprule
\multicolumn{7}{c}{\textbf{Seen Tasks}} \\
\midrule
\textbf{Method} & Seen-A & Seen-B & Seen-C & -- & Avg. \\
\midrule
SmolVLA (Base) & 62 & 48 & 54 & -- & 54.6 \\
\textbf{Ours (JITI) + SmolVLA} & \textbf{80} & \textbf{62} & \textbf{74} & -- & \textbf{72.0} \\
\midrule
\multicolumn{7}{c}{\textbf{Unseen Tasks}} \\
\midrule
\textbf{Method} & Unseen-A & Unseen-B & Unseen-C & Unseen-D & Avg. & \\
\midrule
SmolVLA (Base) & 60 & 54 & 46 & 52 & 53.0 & \\
\textbf{Ours (JITI) + SmolVLA} & \textbf{80} & \textbf{60} & \textbf{62} & \textbf{72} & \textbf{68.6} & \\
\bottomrule
\end{tabular}
}
\endgroup

\end{table}

\subsection{Real-World Validation}
\label{sec:real_world}

To validate the real-world applicability of our JITI framework, we conducted experiments on a physical robotic platform, evaluating its robustness to the latency, uncertainty, and stochasticity inherent in real-world environments.

\noindent \textbf{Real-World Task Suite.}
We designed a suite of six manipulation tasks, each evaluated over 50 rollouts. 
The tasks were selected to assess both physical precision and behavioral smoothness, key aspects of ``elegant'' execution. 
The suite covers common household scenarios, including drawer closing, object placing, and multi-object stacking of varying difficulty.
Detailed task descriptions and evaluation criteria are provided in Supp.~\ref{sec:realworld_setting}.

\noindent \textbf{Experimental Setup.}
We deploy our system on a \textbf{SO-100} robotic arm using \textbf{SmolVLA} as the base policy. 

\begin{figure}[t]
    \centering
    \includegraphics[width=0.48\textwidth]{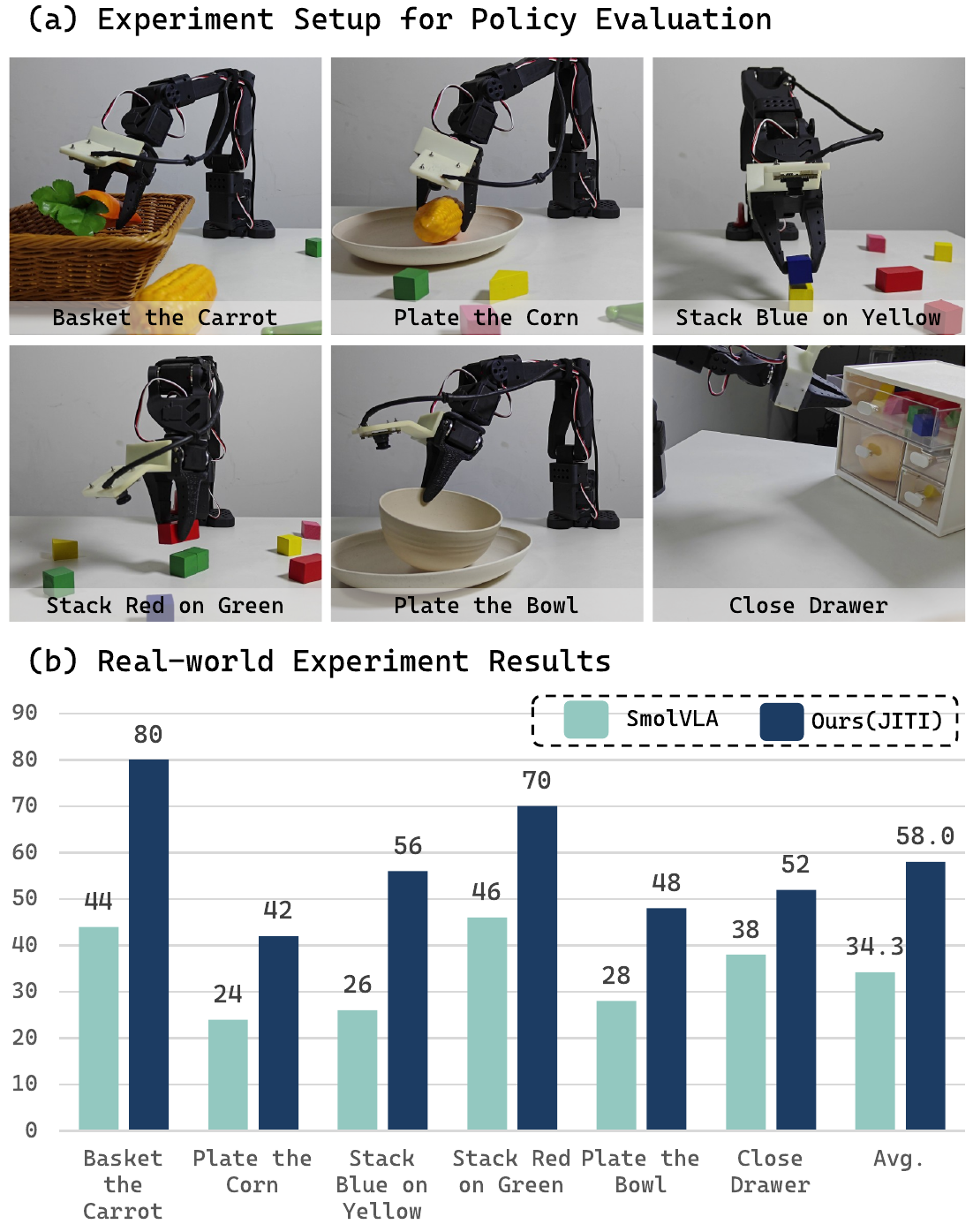}
    \caption{
        \textbf{Real-world validation on our six-task suite.}
        (a) Task setups for the six tasks.
        (b) Elegant Success Rate (ESR~\%) of the \texttt{Base Policy} (SmolVLA) and \texttt{Ours (JITI)}.
    }
    \label{fig:realworld}
\end{figure}

\noindent \textbf{Results.}
Fig.~\ref{fig:realworld} presents the averaged results across all tasks. 
The \texttt{Base Policy} (SmolVLA) achieves an average ESR of 34.3\% across six real-world tasks.  
In contrast, \textbf{\texttt{Ours (JITI)}} improves the average ESR to 58.0\% (+23.7 pts), showing consistent gains across all tasks.  
The largest improvements occur on precision-demanding tasks such as stacking and placing, highlighting JITI’s effectiveness in refining motion quality under real-world uncertainty.

These findings confirm that our framework not only enhances manipulation quality in simulation but also translates effectively to the real world.
By leveraging critic feedback and temporal Q-value fluctuation, the proposed method provides a lightweight, training-free mechanism for improving motion elegance under real-world constraints.

\section{Conclusion}
In this work, we present a decoupled framework for refining elegant robotic manipulation from mixed-quality demonstrations. By combining an \emph{Elegance Critic} with the \emph{Just-in-Time Intervention (JITI)} mechanism, our approach selectively refines actions at decision-critical moments, improving execution quality without retraining the base policy. Extensive experiments on the LIBERO-Elegant benchmark and real-world robotic platforms show substantial gains in Elegant Success Rate (ESR) across diverse tasks and architectures. 
These findings show that implicit execution quality can be learned from demonstrations and applied during control, supporting generalization across varied tasks and promoting efficient and scalable quality-aware, human-like manipulation \emph{beyond success}.




{
    \small
    \bibliographystyle{ieeenat_fullname}
    \bibliography{main}
}

\clearpage
\appendix
\setcounter{page}{1}
\maketitlesupplementary

\section{Overview}
\label{sec:supp_overview}

Considering the space limitation of the main paper, we provide additional details and results in this supplementary appendix. 
Specifically, we present comprehensive descriptions of the \textbf{LIBERO-Elegant Benchmark}, detailed \textbf{implementation settings} for all baselines, and extended \textbf{experiment configurations} in both simulation and real-world environments. 
This appendix is organized as follows:

\begin{enumerate}[label=\Alph*., start=2]
    \item \hyperref[sec:libero_appendix]{\textbf{LIBERO-Elegant Benchmark Details}}
    \begin{enumerate}[label=\arabic*.]
        \item \hyperref[sec:libero_motivation]{Motivation and Construction}
        \item \hyperref[sec:libero_tasklist]{Task List and Structure}
        \item \hyperref[sec:libero_annotation]{Annotation Protocol and Reward Design}
    \end{enumerate}

    \item \hyperref[sec:supp_imple]{\textbf{Implementation Details for RQ1}}
    \begin{enumerate}[label=\arabic*.]
        \item \hyperref[sec:train_pi05]{Training Configuration for $\pi_{0.5}$}
        \item \hyperref[sec:train_gr00t_n1_n15]{Training Configuration for GR00T}
        \item \hyperref[sec:train_smolvla]{Training Configuration for SmolVLA}
        \item \hyperref[sec:jiti_critic_training]{Elegance Critic Training Details}
    \end{enumerate}

    \item \hyperref[sec:supp_rq3]{\textbf{Task Specification for Generalization (RQ3)}}
    \begin{enumerate}[label=\arabic*.]
        \item \hyperref[sec:gen_setup]{Generalization Setup: Seen and Unseen Tasks}
        \item \hyperref[sec:gen_specs]{Task Specifications and Quantitative Analysis}
    \end{enumerate}
    
    \item \hyperref[sec:realworld_setting]{\textbf{Real-World Experiment Suite}}
    \begin{enumerate}[label=\arabic*.]
        \item \hyperref[sec:realworld_envsetup]{Experimental Setup}
        \item \hyperref[sec:realworld_tasksuite]{Real-World Task Suite}
    \end{enumerate}
    
    \item \hyperref[sec:supp_analysis_future]{\textbf{Extended Qualitative Analysis and Future Work}}
    \begin{enumerate}[label=\arabic*.]
        \item \hyperref[sec:case_study_detail]{Case Study: JITI-Guided Elegance Refinement}
        \item \hyperref[sec:future_work]{Future Work}
    \end{enumerate}

    \item \hyperref[sec:release_plan]{\textbf{Benchmark Release Plan}}
\end{enumerate}

    






\section{LIBERO-Elegant Benchmark Details}
\label{sec:libero_appendix}

\subsection{Motivation and Construction}
\label{sec:libero_motivation}

\noindent \textbf{Motivation.}
Existing LIBERO~\cite{liu2023libero} datasets cover a wide range of manipulation tasks and provide rich demonstration data.
However, they evaluate performance only by \textit{whether} the goal is achieved.
In practice, many demonstrations succeed but differ noticeably in how well they satisfy implicit task expectations.
Such variation is currently ignored, making it difficult to measure and improve execution quality~\cite{Mu_2025_CVPR, li2024evaluating}.


\noindent \textbf{Analysis of Existing LIBERO Suites.}
The LIBERO benchmark consists of four task groups, each targeting a distinct generalization dimension:
\begin{itemize}[leftmargin=*]
    \item \textbf{LIBERO-Spatial:} Tests spatial generalization, performing the same task under different spatial layouts.
    \item \textbf{LIBERO-Object:} Tests object generalization, executing the same goal with different target objects.
    \item \textbf{LIBERO-Goal:} Tests goal generalization, covering diverse goal types.
    \item \textbf{LIBERO-100:} Combines all the above variations to evaluate compositional generalization.
\end{itemize}
These suites are well suited for evaluating task completion across diverse conditions, but they still treat all successful executions as equivalent.

\noindent \textbf{Motivated Subset Construction.}
To fill this gap, we introduce \textbf{LIBERO-Elegant}, a curated subset where the \textit{quality of execution} becomes an explicit evaluation dimension.
Tasks in LIBERO-Elegant are chosen because they rely on Implicit Task Constraints (ITCs) that influence behavioral quality, enabling more fine-grained comparison among successful trajectories.
This provides a controlled environment for studying methods that refine policy behavior without changing the underlying goal.

\subsection{Task List and Structure}
\label{sec:libero_tasklist}

\noindent
\textbf{Overview.}
The LIBERO-Elegant benchmark consists of eight manipulation tasks selected for their sensitivity to motion quality and implicit task constraints (ITCs).
Each task inherits the base instruction and success condition from the original LIBERO suite but introduces additional qualitative criteria that emphasize how the task is performed rather than merely whether it succeeds.

\noindent
\textbf{Task Composition.}
To cover a diverse range of execution qualities, the eight tasks are grouped according to four core \textit{Elegance Criteria}:
\begin{itemize}[leftmargin=*]
\item \textbf{Task Sequence Integrity.}
Whether the execution respects the intended ordering and timing of key actions without premature releases, unnecessary pauses, or unintended reversals. 
This dimension ensures that objects remain securely grasped until the correct release moment (Tasks 0–1) and that the overall action sequence progresses smoothly and purposefully toward the goal.

\item \textbf{Target Pose Accuracy.}
Whether the manipulated object reaches the correct final position within a tight spatial tolerance. 
This dimension captures precise placement requirements such as centering a bowl on a plate or positioning a frying pan accurately on a stovetop (Tasks 2–3).

\item \textbf{Pose Alignment.}
Whether the object’s final orientation matches the desired orientation for stable insertion or placement. 
This dimension evaluates rotational correctness, as required when inserting a book with proper orientation into a caddy compartment (Tasks 4–5).

\item \textbf{Collision-Free Execution.}
Whether the trajectory avoids unintended contact with the environment or nearby objects. 
This dimension assesses the safety and spatial awareness of the motion, ensuring smooth pushing without collisions (Task 6) and transporting objects through cluttered spaces without touching neighboring items (Task 7).
\end{itemize}


\noindent
\textbf{Summary.}
This categorization ensures broad manipulation coverage, requiring policies to perform not only successful but also precise, stable, and safe executions across diverse physical constraints.

\noindent
\textbf{Implicit Task Constraints as Concrete Elegance Rules.}
While the Elegance Criteria describe high-level dimensions of execution quality,
the actual evaluation in LIBERO-Elegant is grounded in 
task-specific \emph{Implicit Task Constraints} (ITCs), which define
\emph{how} elegance is judged for each task.
Each task is associated with a primary Elegance Criterion (last column of
Table~\ref{tab:libero_tasklist}), and its corresponding ITC (third column)
specifies a concrete, verifiable rule that reflects this criterion.
For example, ``no premature release before the object is fully inside the
container'' instantiates \emph{Task Sequence Integrity}, while
``centering the bowl on the plate within a tight tolerance'' reflects
\emph{Target Pose Accuracy}.
These ITCs form the basis for selecting critical motion segments and
assigning rewards in our LIBERO-Elegant annotation process. 
Although defined per task, these ITCs naturally generalize across semantically related manipulation behaviors, enabling consistent evaluation beyond the curated eight tasks.

Table~\ref{tab:libero_tasklist} summarizes this mapping from each task’s
behavioral requirement to its corresponding ITC and primary quality
dimension.
Figure~\ref{fig:supp_libero_elegant_tasks} further visualizes elegant versus
non-elegant executions, highlighting where ITC violations occur along the
trajectory.

\noindent
\textbf{Operationalizing ITCs in the LIBERO Simulator.}
In the original LIBERO setup, task success is specified using the
Behavior Domain Definition Language (BDDL), which provides predicates such as
\texttt{In}, \texttt{On}, \texttt{Open}, \texttt{Close}, \texttt{TurnOn}, and
\texttt{TurnOff} to check whether the final goal conditions are satisfied.
However, these predicates encode only \emph{what} outcome is achieved, without
constraining \emph{how} the object is manipulated throughout the trajectory.

To enforce Implicit Task Constraints (ITCs), we extend the BDDL predicate set
with additional semantics that ground the Elegance Criteria in simulation:
\begin{itemize}[leftmargin=*]
\item \texttt{IsGrasping}: ensures that the manipulated object remains securely grasped until the intended release moment.
\item \texttt{IsOnBottomOf}: verifies that the object is stably supported by the target surface without premature dropping.
\item \texttt{IsPreciselyOn}: enforces accurate placement within a tight positional tolerance.
\item \texttt{IsOrientationAligned}: checks rotational correctness required for stable insertion or placement.
\item \texttt{PositionUnchanged}: confirms that nearby objects are not unintentionally disturbed by the motion.
\end{itemize}

These predicates are incorporated into the BDDL goals of each task to produce
an ITC-aware evaluation rule. For instance, Task~0 requires the ketchup bottle
not only to be \texttt{In} the basket region, but also \texttt{IsOnBottomOf}
the basket and still \texttt{IsGrasping} at release, preventing premature
dropping. Tasks~2--3 enforce precise placement via \texttt{IsPreciselyOn}, and
Tasks~6--7 penalize unintended contact through \texttt{PositionUnchanged}.

By embedding ITCs directly into the simulator, LIBERO-Elegant provides
automated, consistent evaluation of execution quality: succeeding at the task
is necessary but \emph{not} sufficient, the execution must also satisfy the
implicit quality constraints that define elegant manipulation.

\subsection{Annotation Protocol and Reward Design}
\label{sec:libero_annotation}

\noindent
\textbf{Motivation.}
While the LIBERO-Elegant benchmark defines the qualitative dimensions of manipulation quality, 
its effective use for learning requires explicit supervision at decision-critical moments. 
Rather than relying solely on sparse, trajectory-level success signals, we adopt a 
\textit{task-specific, segment-level annotation scheme} to capture execution quality with respect to Implicit Task Constraints (ITCs).

\noindent
\textbf{Annotation Procedure.}
For each demonstration, we manually identify one or more short temporal segments where the ITCs are most relevant, 
and assign \textbf{binary rewards} $r_t \in \{0,1\}$ indicating whether the constraint is satisfied. 
For example, in \texttt{Task-5} (\textit{place the book in the right compartment of the caddy}), 
we focus on the final alignment phase: trajectories with precise, controlled placement receive $1$, 
while premature release or misalignment yields $0$. 
Similarly, in \texttt{Task-6} (\textit{push the plate to the front of the stove}), 
reward is given only when the robot avoids contacting surrounding objects during navigation.

\begin{figure*}[t]
    \centering
    \includegraphics[width=\textwidth]{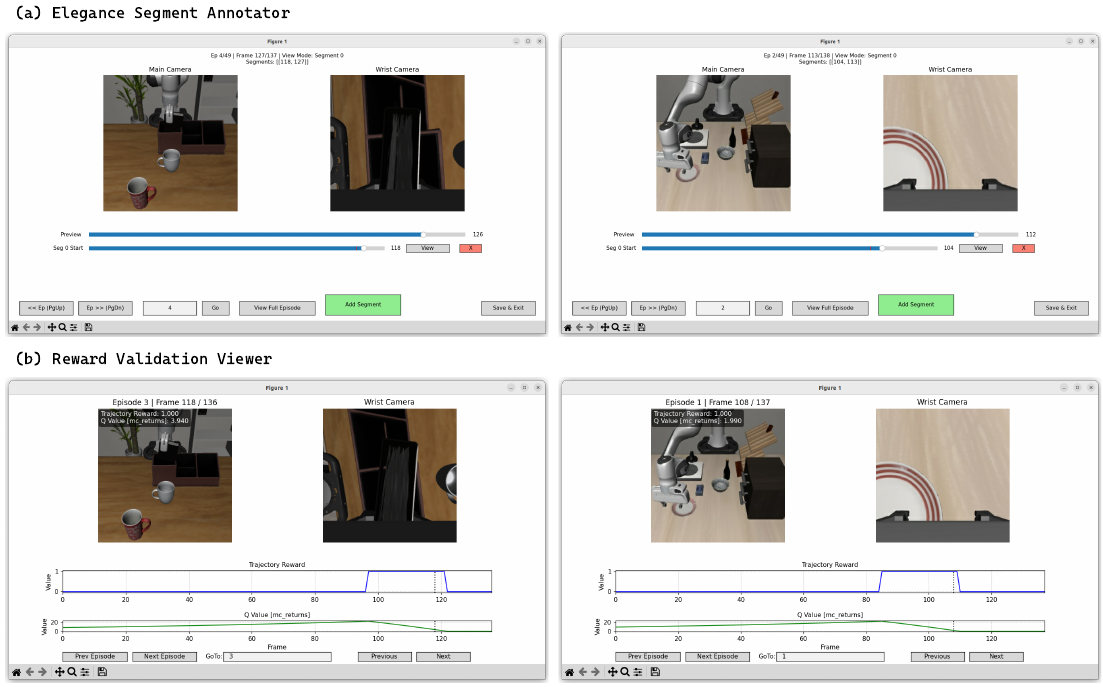}
    \caption{\textbf{Overview of the custom tools developed for our elegance annotation workflow}, 
    exemplified using \texttt{Task-5} (placing the book) and \texttt{Task-6} (pushing the plate). 
    (a) The \textbf{Elegance Segment Annotator (ESA)} enables annotators to interactively select 
    key-motion segments where implicit task constraints (ITCs) are evaluated. 
    (b) The \textbf{Reward Validation Viewer (RVV)} visualizes the annotated per-frame rewards 
    and the resulting Monte Carlo returns, allowing rapid validation of reward quality 
    and internal consistency.}
    \label{fig:annotation_tools}
\end{figure*}

\noindent
\textbf{Annotation Tools.}
To facilitate this process and ensure consistency, we developed two custom visual tools (Figure~\ref{fig:annotation_tools}). 
The \textbf{Elegance Segment Annotator (ESA)} (Figure~\ref{fig:annotation_tools}(a)) enables annotators to load a trajectory, 
inspect multi-view video streams, and interactively mark short evaluation segments at positions where ITCs apply. 
The tool stores the resulting $[start, end]$ indices as annotation metadata for later reward assignment. 
The \textbf{Reward Validation Viewer (RVV)} (Figure~\ref{fig:annotation_tools}(b)) is used for post-annotation verification, 
allowing annotators to review the temporal distribution of binary rewards and ensure correct alignment with constraint-critical motion.


\noindent
\textbf{Outcome.}
This task-specific supervision yields the \textit{Elegance-Enriched Dataset} $\mathcal{D}_{\text{elegant}}$, 
incorporating \textbf{time-aligned} binary indicators of constraint satisfaction throughout the trajectory. 
This high-quality data serves as the critical bedrock for training the elegance-aware value estimator in Stage~2.

\twocolumn[{%
\renewcommand\twocolumn[1][]{#1}%
\begin{center}
    \centering
    \includegraphics[width=1\textwidth]{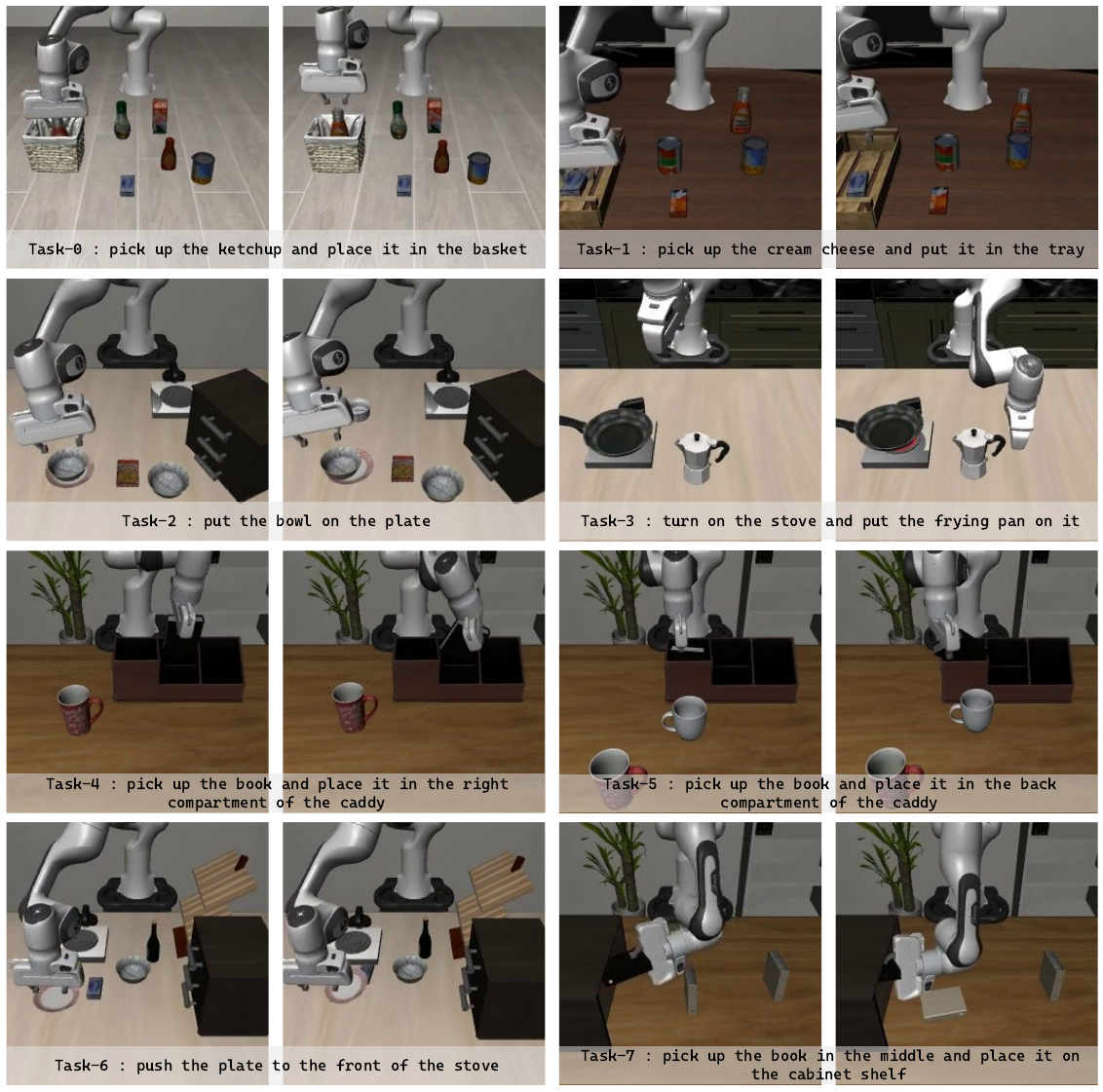}
    \captionof{figure}{\textbf{Visual examples of the LIBERO-Elegant benchmark tasks.} For each task, the left image shows a trajectory satisfying the corresponding \textit{Implicit Task Constraint (ITC)}, while the right image illustrates a non-elegant execution violating that constraint (e.g., early release, misalignment, or unintended collision). This visual contrast highlights the quality dimension emphasized by each \textbf{Elegance Criteria}.}
    \label{fig:supp_libero_elegant_tasks}
\end{center}%
}]

\begin{table*}[t]
\small
\centering
\caption{
\textbf{Task list for the LIBERO-Elegant benchmark.}
Each task inherits a base instruction from LIBERO but introduces a specific \textit{Implicit Task Constraint (ITC)} that targets one of four \textbf{Elegance Criteria}: 
Task Sequence Integrity, Target Pose Accuracy, Pose Alignment, or Collision-Free Execution.
}
\label{tab:libero_tasklist}
\renewcommand{\arraystretch}{1.0}
\setlength{\tabcolsep}{3pt}
\begin{tabular}{c c c c}
\toprule
\textbf{Task ID} & \textbf{Instruction} & \textbf{Implicit Task Constraint (ITC)} & \textbf{Elegance Criteria} \\
\midrule
Task 0 &
\parbox[c][3.0em][c]{5cm}{\centering pick up the ketchup and place it in the basket} &
\parbox[c][3.0em][c]{7cm}{\centering The object must remain securely grasped and only be released once fully inside the basket, ensuring no premature drop.} &
\parbox[c][3.0em][c]{3cm}{\centering Task Sequence Integrity} \\
Task 1 &
\parbox[c][3.0em][c]{5cm}{\centering pick up the cream cheese and put it in the tray} &
\parbox[c][3.0em][c]{7cm}{\centering The gripper must retain the object until it is stably placed in the tray; early release constitutes a failure.} &
\parbox[c][3.0em][c]{3cm}{\centering Task Sequence Integrity} \\
Task 2 &
\parbox[c][3.0em][c]{5cm}{\centering put the bowl on the plate} &
\parbox[c][3.0em][c]{7cm}{\centering The bowl must be placed precisely at the plate’s center, maintaining positional and rotational accuracy.} &
\parbox[c][3.0em][c]{3cm}{\centering Target Pose Accuracy} \\
Task 3 &
\parbox[c][3.0em][c]{5cm}{\centering turn on the stove and put the frying pan on it} &
\parbox[c][3.0em][c]{7cm}{\centering The frying pan must be centered on the stovetop within a small positional tolerance.} &
\parbox[c][3.0em][c]{3cm}{\centering Target Pose Accuracy} \\
Task 4 &
\parbox[c][3.0em][c]{5cm}{\centering pick up the book and place it in the back compartment of the caddy} &
\parbox[c][3.0em][c]{7cm}{\centering The book must be inserted with correct orientation and full depth into the caddy compartment.} &
\parbox[c][3.0em][c]{3cm}{\centering Pose Alignment} \\
Task 5 &
\parbox[c][3.0em][c]{5cm}{\centering pick up the book and place it in the right compartment of the caddy} &
\parbox[c][3.0em][c]{7cm}{\centering The book must be aligned with the caddy slot and inserted without angular deviation.} &
\parbox[c][3.0em][c]{3cm}{\centering Pose Alignment} \\
Task 6 &
\parbox[c][3.0em][c]{5cm}{\centering push the plate to the front of the stove} &
\parbox[c][3.0em][c]{7cm}{\centering The arm must maintain a smooth trajectory while avoiding any collision with other objects.} &
\parbox[c][3.0em][c]{3cm}{\centering Collision-Free Execution} \\
Task 7 &
\parbox[c][3.0em][c]{3cm}{\centering pick up the book in the middle and place it on the cabinet shelf} &
\parbox[c][3.0em][c]{7.5cm}{\centering The arm must move the book safely without contacting neighboring objects during transport.} &
\parbox[c][3.0em][c]{3cm}{\centering Collision-Free Execution} \\
\bottomrule
\end{tabular}
\vspace{-1ex}
\end{table*}

\section{Implementation Details for RQ1}
\label{sec:supp_imple}

\noindent\textbf{Overview.}
To ensure a fair and controlled comparison of execution quality under JITI refinement,
we fine-tune three representative base policies on the LIBERO-Elegant dataset: 
$\pi_{0.5}$~\cite{pi05}, GR00T-N1/N1.5~\cite{gr00tn1_2025}, and SmolVLA-500M~\cite{SmolVLA}. 
All models share the same observation modalities, embodiment configuration, and data formats, 
differing only in model capacity and trainable components. 

We then integrate our Elegance Critic with these base policies during inference to obtain the 
\emph{JITI-guided} variants evaluated in the main results. 
No additional training is applied to the base policies in this stage.

Table~\ref{tab:base_training_summary} summarizes training steps, batch sizes, and compute requirements
for the three base policies. 
Detailed implementation configurations for both the base policies and their JITI-guided versions 
are provided in the following subsections.

\subsection{Training Configuration for $\pi_{0.5}$}
\label{sec:train_pi05}

\noindent\textbf{Model Architecture.}
The $\pi_{0.5}$~\cite{pi05} model is based on a hybrid vision–language–action architecture.
We fine-tune a \textbf{Paligemma-2B} backbone with LoRA adapters for visual–language encoding, while a \textbf{Gemma-300M} head predicts 10-step action sequences per decision step.
Only the LoRA-modified modules and action head are trainable; the rest of the backbone remains frozen.

\noindent\textbf{Dataset and Observations.}
Training uses the curated \textbf{LIBERO-Elegant} subset in the LeRobotLiberoDataConfig format.
Each sample provides synchronized two-view RGB observations (front and wrist cameras), 8D proprioceptive state, the language instruction prompt, and the corresponding 10-step continuous action sequence.
All base policies share identical observation modalities to ensure controlled comparisons.

\noindent\textbf{Optimization Setup.}
We train for \textbf{30k} steps with a batch size of \textbf{8}, using the AdamW optimizer with a gradient clipping norm of \textbf{1.0}. 
The learning rate follows a cosine decay schedule from a peak value of \textbf{5e-5}, with a warm-up of \textbf{10k} steps. 
EMA is \textbf{disabled}, and weights are initialized from the official \texttt{pi05\_base} checkpoint.

\noindent\textbf{Compute and Runtime.}
All experiments use a single NVIDIA RTX 5090 (32GB) with \texttt{bfloat16} precision.
A full run completes in approximately \textbf{16 hours}.
Checkpoints are saved every \textbf{10k} steps.

\subsection{Training Configuration for GR00T-N1 and GR00T-N1.5}
\label{sec:train_gr00t_n1_n15}

\noindent\textbf{Model Architecture.}
We fine-tune the GR00T-N1 (2B)~\cite{gr00tn1_2025} and GR00T-N1.5 (3B)~\cite{gr00tn1_2025} models to serve as additional base policies. 
For both variants, the \emph{vision and language backbones remain frozen}. 
Only the projection layers and the diffusion-based action policy head are updated. 
LoRA tuning is disabled.

\noindent\textbf{Dataset and Observations.}
The models are trained on LIBERO-Elegant using the \texttt{libero\_panda\_gripper} embodiment configuration, 
with identical observation modalities as in $\pi_{0.5}$: two-view RGB images (front and wrist cameras), 
8D proprioceptive inputs, language instruction prompts, and the corresponding continuous action sequences.

\noindent\textbf{Optimization Setup.}
Both experiments follow the official GR00T training framework, with customized hyperparameters adapted for the LIBERO-Elegant benchmark.
Each model is trained for \textbf{50k} steps with a batch size of \textbf{16}. 
We use AdamW with a learning rate of \textbf{1e-4}, a warm-up ratio of \textbf{0.05}, 
and a weight decay of \textbf{1e-5}. 
Checkpoints are written every \textbf{5k} steps.

\noindent\textbf{Compute and Runtime.}
Training is performed on \textbf{4$\times$ NVIDIA A40} GPUs with \texttt{bfloat16} precision. 
A full run takes approximately \textbf{40 hours} to complete 50k steps.

\subsection{Training Configuration for SmolVLA}
\label{sec:train_smolvla}

\noindent\textbf{Model Architecture.}
We fine-tuned the \textbf{SmolVLA (500M)}~\cite{SmolVLA} model on the \textbf{LIBERO-Elegant} dataset to serve as the base policy for our JITI-guided refinement experiments. The model was trained using the official \texttt{lerobot} policy training framework, with a configuration adapted for fine-grained manipulation quality evaluation.

\noindent\textbf{Dataset and Observations.}
Training uses the LIBERO-Elegant dataset with the same observation configuration as other base policies: 
two-view RGB images (front and wrist cameras), 8D proprioceptive inputs, the language instruction prompt, and the corresponding ground-truth action sequence.

\noindent\textbf{Optimization Setup.}
We train for \textbf{100k} steps with a batch size of \textbf{64}, using AdamW 
($\beta_1=0.9, \beta_2=0.95, \epsilon=10^{-8}$) and a learning rate of \textbf{1e-4}. 
The LR follows a cosine decay schedule with \textbf{1k} warm-up steps.
Mixed precision is disabled for stability. 
Checkpoints are saved every \textbf{25k} steps, and evaluations are conducted every \textbf{20k} steps.

\noindent\textbf{Compute and Runtime.}
Training is performed using a single NVIDIA RTX 5090 (32GB GPU). 
A full run takes approximately \textbf{17 hours}.

\begin{table}[h]
\centering
\caption{\textbf{Training summary of base policies evaluated in RQ1.}}
\label{tab:base_training_summary}
\small
\resizebox{\columnwidth}{!}{%
\begin{tabular}{lcccc}
\toprule
\textbf{Model} & \textbf{Steps} & \textbf{Batch} & \textbf{Compute (GPU)} & \textbf{Time} \\
\midrule
$\pi_{0.5}$ & 30k & 8 & RTX~5090~(1) & 16 h \\
GR00T-N1 & 50k & 16 & NVIDIA~A40~(4) & 40 h \\
GR00T-N1.5 & 50k & 16 & NVIDIA~A40~(4) & 40 h \\
SmolVLA-500M & 100k & 64 & RTX~5090~(1) & 17 h \\
\bottomrule
\end{tabular}
}
\end{table}

\subsection{Elegance Critic Training Details}
\label{sec:jiti_critic_training}

\noindent\textbf{Training with SmolVLA Features.} 
For the SmolVLA-based setup, the Elegance Critic is trained on top of \emph{frozen} SmolVLA features. 
The backbone processes two-view RGB observations, proprioceptive states, 
and the language instruction, and we project its hidden representation into the critic backbone space.
The critic predicts values over action chunks of length $K=10$ in the LIBERO simulation environment,
with a matching temporal-difference offset $\text{offset} = 10$.
We train for 20k gradient steps with a batch size of $32$, using a soft target-network update rate 
$\rho = 5.0 \times 10^{-3}$.

The optimization setup uses separate learning rates for different components:
the actor and critic heads use $\text{LR}_{\pi} = \text{LR}_{Q} = 1.0 \times 10^{-5}$, 
the visual--language backbone and projection layers use 
$\text{LR}_{\text{VLM}} = 3.0 \times 10^{-6}$, 
the temperature parameter $\tau$ uses $\text{LR}_{\tau} = 2.0 \times 10^{-5}$, 
and the CQL~\cite{kumar2020conservative} regularization weight $\alpha$ is updated with 
$\text{LR}_{\alpha} = 1.0 \times 10^{-4}$.
We adopt a fixed CQL coefficient $\alpha = 5.0$ without autotuning, 
with a target action gap of $0.5$.
All models are trained in \texttt{bfloat16} with gradient checkpointing enabled.
The critic backbone is based on \texttt{SmolVLM2-256M-Video-Instruct}~\cite{marafioti2025smolvlm}, 
fed by a linear projection from the SmolVLA feature dimension to its hidden size.

\noindent\textbf{Training with GR00T-N1.5 Features.}
For the GR00T-based setup, we follow a similar Cal-QL~\cite{nakamoto2023cal} training scheme, 
but use frozen GR00T-N1.5 features as input. 
GR00T-N1.5 encodes the same observation tuple, and all GR00T components remain frozen during critic training.
The critic operates on action chunks of length $K=10$ with base action dimension $D_{\text{act}} = 7$.
We use $\gamma = 0.98$, batch size $32$, and train for 20k gradient steps, with gradient norm clipping at $0.5$ and the same soft update rate $\rho = 5.0 \times 10^{-3}$.

The learning rate configuration mirrors the SmolVLA case, 
except for a lower actor learning rate $\text{LR}_{\pi} = 1.0 \times 10^{-6}$ 
to stabilize training under the diffusion-based action space. 
We again use a fixed CQL coefficient $\alpha = 5.0$ without autotuning, 
with a target action gap of $0.5$ and an (unused) Lagrange multiplier 
initialized at $0.1$. 
The critic backbone here is \texttt{nvidia/Eagle2-1B}~\cite{li2025eagle}, 
with a linear projection from the GR00T feature dimension to the 
corresponding hidden size, and all components are trained in 
\texttt{bfloat16} precision.

\section{Task Specification for Generalization (RQ3)}
\label{sec:supp_rq3}

\noindent
\textbf{Overview.}
This section provides detailed task definitions and configurations used in the generalization study (RQ3), where the Elegance Critic trained on a limited set of tasks is evaluated on unseen but semantically related tasks.

\subsection{Generalization Setup: Seen and Unseen Tasks}
\label{sec:gen_setup}
\begin{figure*}[t]
    \centering
    \includegraphics[width=\textwidth]{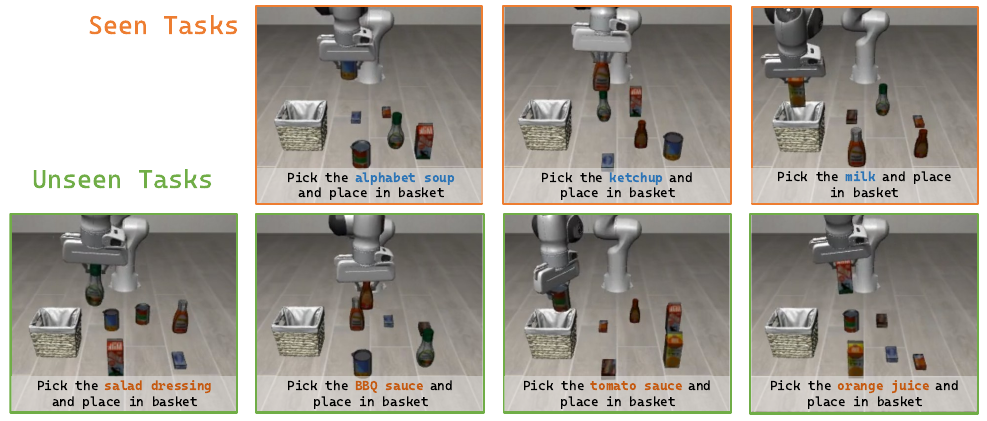}
    \caption{\textbf{Visualization of tasks used in the generalization study.}
    The top row shows the three \textbf{Seen Tasks} (\textit{alphabet soup}, \textit{ketchup}, and \textit{milk}) 
    used for training the Elegance Critic, while the bottom row shows the four \textbf{Unseen Tasks} 
    (\textit{salad dressing}, \textit{BBQ sauce}, \textit{tomato sauce}, and \textit{orange juice}) used for 
    evaluation without fine-tuning. All tasks share identical success conditions but differ in spatial layouts, 
    object geometry, and visual appearance.}
    \label{fig:supp_gen_tasks}
\end{figure*}

To assess the critic’s ability to generalize its learned notion of elegance, 
we evaluate it on a family of pick-and-place tasks divided into two groups, 
as illustrated in Figure~\ref{fig:supp_gen_tasks}.
Specifically:
\begin{itemize}[leftmargin=*]
    \item \textbf{Seen Tasks:} 
    Three tasks used for critic training, placing everyday objects such as \textit{milk}, \textit{ketchup}, and \textit{alphabet soup} into a basket.
    These tasks share identical success conditions but differ in spatial layouts, object geometry and visual appearance. Additionally, these tasks require that the object remain securely grasped throughout the entire motion and only be released once fully inside the basket, ensuring no premature drops.
    \item \textbf{Unseen Tasks:} 
    Four novel but semantically similar tasks, placing \textit{salad dressing}, \textit{BBQ sauce}, \textit{tomato sauce}, and \textit{orange juice} into the same basket.
    These tasks are excluded during training to test the critic’s ability to generalize to unseen object appearances and spatial layouts. Similar success conditions apply, enforcing secure grasping and proper release timing to meet criteria for elegant task execution.
\end{itemize}

\subsection{Task Specifications and Quantitative Analysis}
\label{sec:gen_specs}
Table~\ref{tab:gen_tasklist} provides detailed specifications of these tasks, including instructions and elegance criteria. Additionally, quantitative evaluations in the main paper show substantial ESR gains on both seen and unseen tasks, demonstrating strong transferability.


\section{Real-World Experiment Suite}
\label{sec:realworld_setting}

\noindent
\textbf{Overview.}
We validated our approach on a physical robotic system to assess its real-world practicality. These tests evaluate whether the Elegance Critic can effectively enhance policy execution under real-world challenges such as sensor noise, actuation delays, and object variability, conditions that are difficult to replicate in simulation.

\subsection{Experimental Setup}
\label{sec:realworld_envsetup}
\noindent \textbf{Hardware.}
Our experimental setup consists of a \textbf{SO100} arm equipped with its native gripper. Visual observations are captured by two cameras : a static front camera and a wrist-mounted camera, both providing 640$\times$480 RGB data. All policy inference and control commands are processed on a desktop workstation with an \textbf{NVIDIA RTX 5090 GPU}.

\subsection{Real-World Task Suite}
\label{sec:realworld_tasksuite}
We designed a suite of 6 real-world manipulation tasks, derived from common household activities. As detailed in Table \ref{tab:real_tasklist}, each task is defined not only by its base instruction (e.g., ``put the bowl on the plate'') but by a corresponding \textit{Implicit Task Constraint (ITC)} that specifies the quality of the execution.

\begin{figure*}[t]
    \centering
    \includegraphics[width=\textwidth]{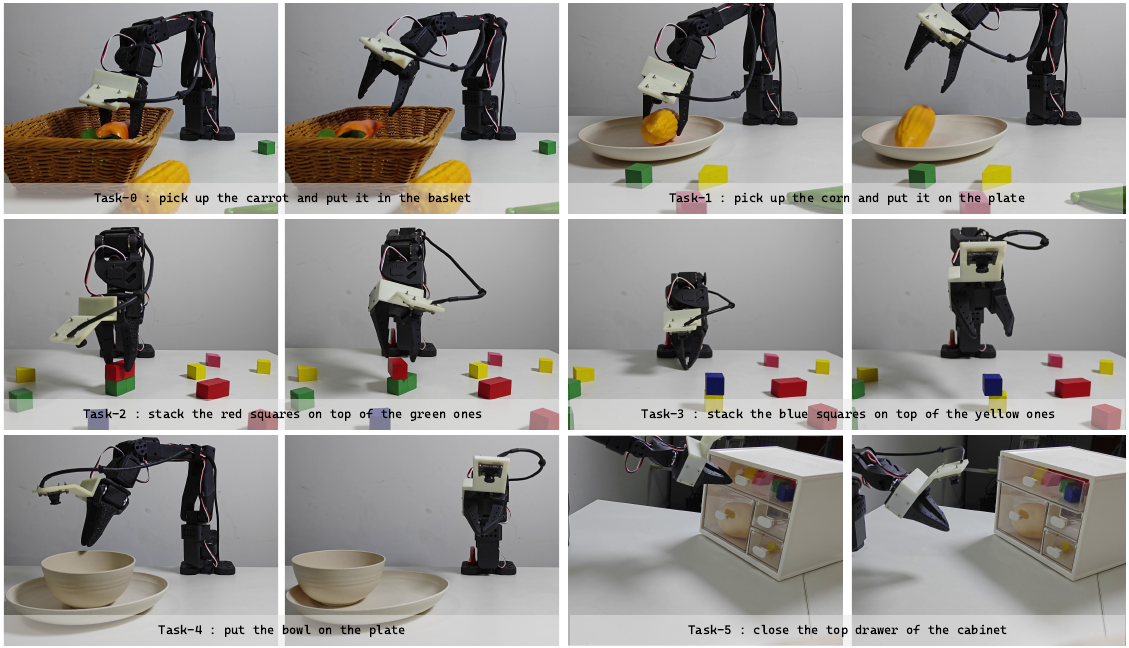}
    \caption{\textbf{Visual overview of the real-world task suite (defined in Table \ref{tab:real_tasklist}).} 
    For each task, we present a side-by-side comparison: 
    (\textbf{Left}) An elegant execution that successfully satisfies the task's \textit{Implicit Task Constraint (ITC)}. 
    (\textbf{Right}) A non-elegant execution that violates the constraint, resulting in failures such as premature dropping, 
    poor alignment, or incomplete motion. This comparison visualizes the fine-grained \textbf{Elegance Criteria} we evaluate.}
    \label{fig:supp_real_tasks}
\end{figure*}

To visually illustrate these constraints, Figure \ref{fig:supp_real_tasks} presents a side-by-side comparison for each task, contrasting a successful, elegant execution against a non-elegant failure case.

\noindent \textbf{Software and Control}
The system is built entirely within the \textbf{LeRobot software framework}. We implement our JITI-guided \textbf{SmolVLA} policy within this framework. The control loop runs at \textbf{30 Hz}. At each timestep, the system receives RGB observations from both cameras, the JITI framework triggers intervention if necessary, and the resulting action dictionary is sent to the robot's low-level controller.

\section{Extended Qualitative Analysis and Future Work}
\label{sec:supp_analysis_future}

\noindent
\textbf{Overview.}
In this section, we provide a detailed qualitative case study to illustrate the fine-grained decision-making of our model, followed by a discussion on limitations and potential avenues for future research.

\subsection{Case Study: JITI-Guided Elegance Refinement}
\label{sec:case_study_detail}
To further illustrate how JITI refines actions at decision-critical states, we present a qualitative case study conducted on a shelf-placement task.

At a decision-critical moment, the Elegance Critic evaluates eight candidate action sequences sampled from the base VLA policy. 
We then branch the rollout into three distinct trajectories: one that
always selects the \textbf{Best-Q} candidate with the highest predicted
elegance, another that selects the \textbf{Median-Q} candidate ranked in
the middle, and a third that selects the \textbf{Worst-Q} candidate with
the lowest predicted elegance.

Each trajectory continues to apply the same selection rule at every
JITI intervention trigger, always selecting the highest, middle, or
lowest-ranked candidate respectively. This causes the three behaviors
to diverge over time, demonstrating how persistent high- or low-quality
decisions ultimately translate into substantial differences in execution
elegance.

\begin{figure*}[t]
    \centering
    \includegraphics[width=\textwidth]{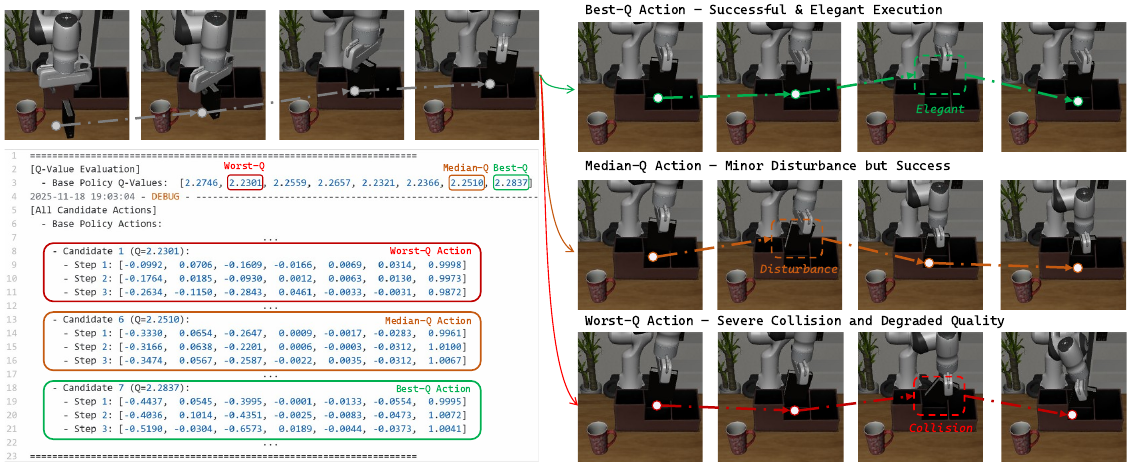}
    \caption{\textbf{Effect of JITI on execution elegance at a decision-critical moment.}
    We visualize three rollout branches using Best-Q, Median-Q, and Worst-Q intervention strategies 
    at every JITI trigger. Higher predicted Q-values result in smoother, collision-free execution
    and correct pose alignment, whereas repeated low-Q selection leads to severe collisions 
    and degraded object orientation.}
    \label{fig:case_jiti}
\end{figure*}

As visualized in Figure~\ref{fig:case_jiti}, the Best-Q trajectory maintains smooth, collision-free motion and achieves precise placement of the book into the target compartment with correct pose alignment. The Median-Q trajectory incurs a slight edge contact and a small amount of tilt, but still results in successful placement. In contrast, the Worst-Q trajectory suffers from repeated collisions with the shelf and severely degraded object orientation, despite eventually completing the task.




\subsection{Future Work}
\label{sec:future_work}

Future directions include leveraging VLMs to infer open-ended elegance requirements from language and enabling the Elegance Critic to actively refine action generation rather than merely selecting among discrete candidates. These extensions will further close the gap between functional success and human-like dexterity in shared environments.

\section{Benchmark Release Plan}
\label{sec:release_plan}
We commit to fully releasing the \textbf{LIBERO-Elegant Benchmark} upon publication, including all demonstrations, segment-level elegance annotations, annotation tools, and evaluation scripts. To promote reproducibility and community adoption, the dataset will be hosted on a public repository with detailed documentation on task setups, annotation protocols, and elegance evaluation criteria.

\begin{table*}[t]
\small
\centering
\caption{
\textbf{Task specification for the generalization study (RQ3).}
The Elegance Critic is trained on three \textbf{Seen Tasks} and evaluated directly on four \textbf{Unseen Tasks} without fine-tuning.
All tasks share the same pick-and-place semantics but differ in object.
}
\label{tab:gen_tasklist}
\renewcommand{\arraystretch}{1.0}
\setlength{\tabcolsep}{3pt}
\begin{tabular}{c c c c}
\toprule
\textbf{Category} & \textbf{Object} & \textbf{Instruction} & \textbf{Elegance Criteria} \\
\midrule
\multicolumn{4}{c}{\textbf{Seen Tasks (Training)}} \\
\midrule
Seen-A & \textit{milk} & \parbox[c][2.0em][c]{7cm}{\centering pick up the milk and place it in the basket} & Task Sequence Integrity \\
Seen-B & \textit{ketchup} & \parbox[c][2.0em][c]{7cm}{\centering pick up the ketchup and place it in the basket} & Task Sequence Integrity \\
Seen-C & \textit{alphabet soup} & \parbox[c][2.0em][c]{7cm}{\centering pick up the alphabet soup and place it in the basket} & Task Sequence Integrity \\
\midrule
\multicolumn{4}{c}{\textbf{Unseen Tasks (Evaluation)}} \\
\midrule
Unseen-A & \textit{salad dressing} & \parbox[c][2.0em][c]{7cm}{\centering pick up the salad dressing and place it in the basket} & Task Sequence Integrity \\
Unseen-B & \textit{BBQ sauce} & \parbox[c][2.0em][c]{7cm}{\centering pick up the BBQ sauce and place it in the basket} & Task Sequence Integrity \\
Unseen-C & \textit{tomato sauce} & \parbox[c][2.0em][c]{7cm}{\centering pick up the tomato sauce and place it in the basket} & Task Sequence Integrity \\
Unseen-D & \textit{orange juice} & \parbox[c][2.0em][c]{7cm}{\centering pick up the orange juice and place it in the basket} & Task Sequence Integrity \\
\bottomrule
\end{tabular}
\vspace{-1ex}
\end{table*}

\begin{table*}[t]
\small
\centering
\caption{
\textbf{Task list for the real-world evaluation suite.}
Each task inherits a base instruction but introduces a concrete \textit{Implicit Task Constraint (ITC)} that emphasizes motion quality and defines the corresponding Elegance Criteria.
}
\label{tab:real_tasklist}
\renewcommand{\arraystretch}{1.0}
\setlength{\tabcolsep}{3pt}
\begin{tabular}{c c c c}
\toprule
\textbf{Task ID} & \textbf{Instruction} & \textbf{Implicit Task Constraint (ITC)} & \textbf{Elegance Criteria} \\
\midrule
Task 0 &
\parbox[c][3.0em][c]{5cm}{\centering pick up the carrot and put it in the basket} &
\parbox[c][3.5em][c]{7cm}{\centering The gripper must hold the carrot firmly and release only after it is fully inside the basket to avoid premature dropping.} &
\parbox[c][3.0em][c]{3cm}{\centering Task Sequence Integrity} \\
Task 1 &
\parbox[c][3.0em][c]{5cm}{\centering pick up the corn and put it on the plate} &
\parbox[c][3.5em][c]{7cm}{\centering The corn must remain stably grasped and released only when securely positioned on the plate, avoiding free fall.} &
\parbox[c][3.0em][c]{3cm}{\centering Task Sequence Integrity} \\
Task 2 &
\parbox[c][3.0em][c]{5cm}{\centering stack the red squares on top of the green ones} &
\parbox[c][3.5em][c]{7cm}{\centering The red block must be placed precisely aligned with the green one, maintaining both positional and rotational accuracy.} &
\parbox[c][3.0em][c]{3cm}{\centering Target Pose Accuracy} \\
Task 3 &
\parbox[c][3.0em][c]{5cm}{\centering stack the blue squares on top of the yellow ones} &
\parbox[c][3.5em][c]{7cm}{\centering The blue block must be placed directly above the yellow block, ensuring stable and centered stacking.} &
\parbox[c][3.0em][c]{3cm}{\centering Target Pose Accuracy} \\
Task 4 &
\parbox[c][3.0em][c]{5cm}{\centering put the bowl on the plate} &
\parbox[c][3.5em][c]{7cm}{\centering The bowl must be centered precisely on the plate with minimal offset or tilt.} &
\parbox[c][3.0em][c]{3cm}{\centering Target Pose Accuracy} \\
Task 5 &
\parbox[c][3.0em][c]{5cm}{\centering close the top drawer of the cabinet} &
\parbox[c][3.5em][c]{7.5cm}{\centering The drawer must be closed smoothly and fully until it is tightly shut.} &
\parbox[c][3.0em][c]{3cm}{\centering Pose Alignment} \\
\bottomrule
\end{tabular}
\vspace{-1ex}
\end{table*}

\end{document}